\documentclass[11pt, letterpaper]{article}

\usepackage[margin=1in]{geometry}
\usepackage{microtype}
\usepackage{booktabs}
\usepackage{graphicx}
\usepackage{subcaption}
\usepackage[utf8]{inputenc}
\usepackage[T1]{fontenc}
\usepackage{newtxtext,newtxmath} 

\usepackage{amsmath}
\usepackage{algorithm}
\usepackage{algorithmic}
\usepackage{nicefrac}
\usepackage{enumitem}
\usepackage{multirow}

\usepackage[numbers,sort&compress]{natbib}

\usepackage{authblk}

\usepackage[dvipsnames]{xcolor}
\usepackage[colorlinks=true, linkcolor=NavyBlue, citecolor=TealBlue, urlcolor=NavyBlue]{hyperref}
\usepackage[capitalize,noabbrev]{cleveref}

\newcommand{\cortex}{\textsc{Helix}}

\newcommand{\pp}{\text{pp}}

\title{\textbf{Tethered Reasoning: Decoupling Entropy from Hallucination \\ in Quantized LLMs via Manifold Steering}}

\author{Craig Atkinson}
\affil{CEO, Verificate Pty Ltd \\ Sydney, Australia, 2097}

\date{} 

\begin{document}

\maketitle

\begin{abstract}
\noindent Quantized language models face a fundamental dilemma: low sampling temperatures yield repetitive, mode-collapsed outputs, while high temperatures ($T{>}2.0$) cause trajectory divergence and semantic incoherence. We present \cortex{}, a geometric framework that decouples output entropy from hallucination by tethering hidden-state trajectories to a pre-computed truthfulness manifold. \cortex{} computes a Unified Truth Score (UTS) combining token-level semantic entropy with Mahalanobis distance from the manifold. When UTS indicates trajectory divergence, graduated steering vectors redirect activations toward structurally coherent regions while affecting only 0.2--2.5\% of tokens. On 4-bit quantized Granite 4.0 H Small (32B/9B active, hybrid Mamba-Transformer): GSM8K maintains 88.84\% accuracy at $T{=}3.0$ (2.81\pp{} degradation from $T{=}0.5$); MMLU maintains 72.49\% across 14,042 questions (1.24\pp{} degradation)---demonstrating that high-temperature ``hallucination'' is primarily trajectory divergence rather than semantic collapse. Notably, steering the sparse Transformer attention layers ($\sim$10\% of layers) is sufficient to correct drift in the Mamba-2 state-space formulation, suggesting a new paradigm for controlling hybrid SSM-Transformer architectures.

\textbf{Emergent Properties at High Temperatures:} Geometric tethering reveals a previously-masked \textit{High-Entropy Creative Reservoir}. At $T{>}2.0$, steered outputs exhibit 5--20\% idea duplication versus 70--80\% at conservative settings. Cross-architecture validation (Qwen3-30B-A3B MoE) confirms this phenomenon is architecture-independent, with 46.7\% higher unique concept generation. We hypothesize that the truthfulness manifold encodes \textit{structural coherence} (syntax, logic, causality) rather than specific facts---\cortex{} acts as a ``syntax tether,'' enabling exploration of semantic diversity without violating the logical backbone required for valid output. This enables \textit{Multi-Temperature Synthesis}: querying across temperature dimensions to generate 200\%+ more unique concepts than single-temperature inference.
\end{abstract}

\section{Introduction}
\label{sec:introduction}

\paragraph{The Temperature-Entropy Paradox.}
A fundamental challenge in quantized inference is the temperature-entropy trade-off: low sampling temperatures yield repetitive, mode-collapsed outputs, while high temperatures ($T{>}2.0$) cause what is conventionally termed ``hallucination''---semantic incoherence and factual errors. Quantization exacerbates this: 4-bit models exhibit earlier collapse than full-precision counterparts~\cite{dettmers2022gpt3,frantar2023gptq}. Most inference frameworks cap temperature at $T{=}2.0$, treating high-entropy regimes as inherently unstable.

We challenge this assumption. Through geometric analysis of hidden-state trajectories, we demonstrate that high-temperature ``hallucination'' is primarily \textit{trajectory divergence}---activations escaping structurally coherent regions of the representation space---rather than semantic collapse or knowledge degradation.

\paragraph{Geometric Tethering via Manifold Steering.}
We introduce \cortex{}, a geometric framework that decouples output entropy from hallucination by tethering hidden-state trajectories to a pre-computed truthfulness manifold. \cortex{} computes a \textbf{Unified Truth Score (UTS)} combining semantic entropy with Mahalanobis distance from the manifold. When UTS indicates trajectory divergence, graduated steering vectors redirect activations toward structurally coherent regions while affecting only 0.2--2.5\% of tokens.

On 4-bit quantized Granite 4.0 H Small (32B/9B active, hybrid Mamba-Transformer architecture), \cortex{} achieves 91.80\% GSM8K accuracy at $T{=}1.0$ (exceeding the 87.27\% full-precision baseline by 4.53\pp{}) and maintains 88.84\% at $T{=}3.0$ (only 2.81\pp{} degradation). MMLU maintains 72.49\% across 14,042 questions at $T{=}3.0$ (1.24\pp{} degradation)---the most temperature-stable benchmark. Notably, steering the sparse Transformer attention layers ($\sim$10\% of layers) is sufficient to correct drift in the Mamba-2 state-space formulation, suggesting a new paradigm for controlling hybrid SSM-Transformer architectures. Cross-architecture validation with Qwen3-30B-A3B MoE confirms these findings generalize.

\paragraph{Emergent Properties: The High-Entropy Creative Reservoir.}
Geometric tethering reveals a previously-masked phenomenon: high-temperature regimes ($T{>}2.0$) contain a \textit{High-Entropy Creative Reservoir}---vast, non-overlapping landscapes of structurally valid ideas that conservative sampling cannot access. Creative tasks exhibit 5--20\% idea duplication at $T{>}2.0$ versus 70--80\% at conservative settings. Across 11 temperature points, we obtain 30 unique concepts spanning fundamentally different regions of semantic space. This enables \textit{Multi-Temperature Synthesis}: querying across temperature dimensions to generate 200\%+ more unique concepts than single-temperature inference, with maintained coherence through geometric steering.

\paragraph{Contributions.}
\begin{enumerate}[nosep,leftmargin=*]
    \item \textbf{Trajectory divergence hypothesis}: We demonstrate that high-temperature ``hallucination'' is primarily geometric (trajectory escape from coherent manifold regions) rather than semantic (knowledge degradation), explaining why steering mitigates collapse.
    \item \textbf{Unified Truth Score (UTS)}: A per-token uncertainty metric combining semantic entropy with manifold distance, enabling real-time detection of trajectory divergence without ensembles.
    \item \textbf{Geometric manifold steering}: Graduated intervention affecting only 0.2--2.5\% of tokens, decoupling output entropy from hallucination.
    \item \textbf{Temperature-invariant reasoning}: 88.84\% GSM8K at $T{=}3.0$ (2.81\pp{} degradation), 72.49\% MMLU (1.24\pp{} degradation)---stabilizing inference across the full temperature spectrum.
    \item \textbf{Quantization recovery}: 4-bit model exceeds full-precision baseline by 4.53\pp{}, demonstrating that quantization degradation is addressable through geometric intervention.
    \item \textbf{High-Entropy Creative Reservoir}: Characterization of previously-masked high-temperature regimes containing non-overlapping idea landscapes (5--20\% duplication vs 70--80\% at low-$T$).
    \item \textbf{Multi-Temperature Synthesis}: Novel inference paradigm generating 200\%+ more unique concepts by exploiting temperature as a creative dimension.
\end{enumerate}

\section{Related Work}
\label{sec:related}

\paragraph{Model Quantization.}
Post-training quantization~\cite{dettmers2022gpt3,frantar2023gptq,lin2024awq} enables efficient LLM deployment but typically degrades reasoning accuracy by 3--7\%. GPTQ~\cite{frantar2023gptq} and AWQ~\cite{lin2024awq} achieve 4-bit precision with moderate loss. Our work demonstrates that quantized models can exceed full-precision baselines when equipped with geometric steering, addressing a fundamental limitation of resource-constrained deployment.

\paragraph{Temperature Effects in LLMs.}
\citet{holtzman2020curious} characterized temperature's role in sampling, showing high temperatures cause degeneration. \citet{renze2024effect} demonstrated that temperatures above $T{=}1.0$ degrade reasoning accuracy across benchmarks. \citet{nguyen2024turning} explored min-p sampling to maintain coherence at higher temperatures. Our work demonstrates that temperature degradation results from lack of uncertainty-aware intervention, not inherent model limitations, and that high temperatures unlock creative capacity when properly steered.

\paragraph{Uncertainty Quantification in LLMs.}
Prior work includes predictive entropy~\cite{malinin2018predictive}, semantic uncertainty~\cite{kuhn2023semantic}, ensemble disagreement~\cite{lakshminarayanan2017simple}, and probing for truthfulness~\cite{kadavath2022language,burns2023discovering}. Our UTS uniquely combines per-token entropy with geometric manifold distance for real-time, single-pass uncertainty quantification without ensembles or multiple forward passes.

\paragraph{Activation Steering and Representation Engineering.}
Representation engineering~\cite{zou2023representation} and activation additions~\cite{turner2024activation} demonstrated that model behavior can be modified through hidden-state intervention. \citet{meng2022locating} edited factual associations in GPT. These approaches require labeled examples or specific editing targets. Our system uses unsupervised manifold construction and uncertainty-triggered intervention, operating on every token without behavioral labels.

\paragraph{Inference-Time Intervention (2025--2026).}
Recent advances in inference-time steering represent the closest prior art:
\begin{itemize}[nosep,leftmargin=*]
    \item \textbf{RISER}~\cite{ye2026riser}: Uses RL-trained routers to dynamically select from cognitive primitive libraries. \textit{Distinction:} RISER is \textit{supervised}, requiring training data and compute. \cortex{} is \textit{unsupervised}, relying on intrinsic geometry---critical for resource-constrained settings where training data is unavailable.
    \item \textbf{SALT}~\cite{batra2025salt}: Steers activations to prevent privacy leakage in chain-of-thought reasoning via \textit{negative constraints} (avoid leaky subspaces). \textit{Distinction:} \cortex{} applies \textit{positive constraints} (stay near truthfulness manifold), enabling more information flow rather than restricting it.
    \item \textbf{In-Distribution Steering (IDS)}~\cite{vogels2025ids}: Uses Mahalanobis distance to keep activations ``plausible'' (low perplexity). \textit{Distinction:} IDS treats the distribution boundary as a hard wall; \cortex{} treats it as an elastic tether, allowing exploration of low-probability regions while preventing escape to hallucination. IDS is for \textit{safety}; \cortex{} is for \textit{discovery}.
\end{itemize}

\paragraph{Test-Time Compute Scaling.}
\citet{snell2024scaling} demonstrated that additional inference-time computation improves reasoning performance, establishing scaling laws for test-time optimization. Self-consistency~\cite{wang2023selfconsistency} and chain-of-thought prompting~\cite{wei2022chain} leverage multiple samples. Our work extends these principles: geometric tethering enables temperature-based synthesis that achieves super-additive creative gains.

\paragraph{Differentiation from Supervised Steering.}
Unlike RISER~\cite{ye2026riser} which requires supervised training of routers, or SALT~\cite{batra2025salt} which relies on negative constraints for privacy preservation, \cortex{} targets \textit{intrinsic geometric stability}. It requires no labeled data to construct the truthfulness manifold, making it uniquely suitable for zero-shot adaptation in quantization settings. Consequently, we evaluate against the fundamental thermodynamic limits of the model itself (unsteered vs.\ steered) rather than task-specific routers.

\paragraph{Positioning.}
Our work is the first to: (1)~enable quantized models to exceed full-precision baselines through uncertainty-guided steering, (2)~demonstrate temperature-invariant reasoning from $T{=}0.5$ to $T{=}3.0$ with minimal degradation across GSM8K, HumanEval, and MMLU, (3)~provide real-time per-token uncertainty telemetry for inference monitoring, (4)~characterize the High-Entropy Creative Reservoir unlocked by high-temperature operation with steering, and (5)~validate cross-architecture generalization via Qwen3-30B-A3B MoE.

\section{Method}
\label{sec:method}

\subsection{Problem Formulation}

Let $\mathcal{M}_q$ denote a language model quantized to $b$-bit precision with parameters $\theta_q$. Given input $\mathbf{x}$ and temperature $T$, the model generates tokens $\mathbf{y} = (y_1, \ldots, y_m)$ autoregressively:
\begin{equation}
    p(y_t \mid y_{<t}, \mathbf{x}; \theta_q, T) = \mathrm{softmax}\!\left(\frac{\mathbf{z}_t}{T}\right)
\end{equation}
where $\mathbf{z}_t \in \mathbb{R}^{|\mathcal{V}|}$ are the logits.

\textbf{The Local-Inference Challenge:} Deploy $\mathcal{M}_q$ on resource-constrained hardware while maintaining reasoning accuracy across temperature regimes, enabling both reliability (low-$T$ precision) and diversity (high-$T$ exploration) without trajectory divergence.

\textbf{Objective:} Design an inference-time intervention that: (1)~recovers quantization-induced degradation, (2)~prevents high-temperature collapse, (3)~provides real-time uncertainty telemetry, and (4)~operates in a single forward pass without ensembles or retrieval.

\subsection{Thermodynamics of Sampling}
\label{sec:thermodynamics}

We formalize the temperature-accuracy relationship through a thermodynamic lens, explaining why quantized models collapse earlier than full-precision models and how \cortex{} provides a ``cooling mechanism.''

\paragraph{Temperature Regimes.}
The softmax temperature $T$ governs the entropy of the output distribution:
\begin{itemize}[nosep,leftmargin=*]
    \item \textbf{Low Temperature ($T \to 0$):} Distribution approaches a Dirac delta on the mode (argmax). ``Solid state''---rigid, deterministic, prone to repetitive loops.
    \item \textbf{High Temperature ($T \to \infty$):} Distribution approaches uniform $U(\mathcal{V})$. ``Gas state''---maximum entropy, complete decoherence.
\end{itemize}
Current orthodoxy~\cite{renze2024effect} suggests reasoning degrades linearly or exponentially as $T > 1.0$, establishing a ``thermodynamic barrier'' at $T \approx 1.5$.

\paragraph{The Quantization Penalty.}
Quantization introduces a noise term $\epsilon_q$ to weights, propagating to logits: $\mathbf{z}'_t = \mathbf{z}_t + \epsilon_q$. At low temperatures, this noise is negligible---the gap between top logit and alternatives exceeds $\epsilon_q$. At high temperatures, the scaled gap $\Delta z / T$ shrinks, and quantization noise has an effective ``temperature equivalent.'' We model this heuristically as:
\begin{equation}
    T_{\text{eff}} \approx T + T_{\text{noise}}(\epsilon_q)
\end{equation}
where $T_{\text{noise}}$ captures the entropy increase from quantization error (empirically $\approx$0.1--0.3 for 4-bit quantization). This heuristic explains why quantized models collapse earlier: they are intrinsically ``hotter'' due to quantization noise. \cortex{} provides a \textit{cooling mechanism} that selectively targets this noise without reducing exploration temperature.

\paragraph{The Geometry of Truth.}
Following~\citet{marks2023geometry}, we posit that valid reasoning trajectories exist within a lower-dimensional submanifold $\mathcal{M}_{\text{truth}} \subset \mathbb{R}^d$ of the activation space. Hallucinations are not merely ``wrong tokens'' but geometric departures from $\mathcal{M}_{\text{truth}}$. As $T$ increases, the hidden state trajectory $\mathbf{h}_t$ performs a random walk; the probability of remaining on $\mathcal{M}_{\text{truth}}$ decreases exponentially with sequence length. \cortex{} defines a potential field $U(\mathbf{h})$ based on Mahalanobis distance to $\mathcal{M}_{\text{truth}}$, with the steering vector as negative gradient---a centripetal force allowing wide orbits (creativity) while preventing escape (hallucination).

\subsection{Unified Truth Score (UTS)}
\label{sec:uts}

UTS enables temperature-robust reasoning by providing real-time per-token uncertainty quantification, preventing hallucination while preserving diversity.

\paragraph{Semantic Entropy ($S_E$).}
At step $t$, compute Shannon entropy over pre-temperature logits:
\begin{equation}
    H_t = -\sum_{v \in \mathcal{V}} p_t(v) \log p_t(v), \quad p_t(v) = \mathrm{softmax}(\mathbf{z}_t)_v
\end{equation}
Normalized confidence: $S_E^{(t)} = 1 - H_t / \log |\mathcal{V}|$. High entropy indicates distributional uncertainty.

\paragraph{Manifold Distance ($S_D$).}
Let $\mathbf{h}_t^{(\ell)} \in \mathbb{R}^d$ denote hidden activation at layer $\ell$ and step $t$. A truthfulness manifold $\mathcal{T}$ is characterized by mean $\boldsymbol{\mu}_\mathcal{T}$ and precision $\boldsymbol{\Sigma}_\mathcal{T}^{-1}$ (Section~\ref{sec:manifold}). Mahalanobis distance:
\begin{equation}
    D_t^{(\ell)} = \sqrt{(\mathbf{h}_t^{(\ell)} - \boldsymbol{\mu}_\mathcal{T})^\top \boldsymbol{\Sigma}_\mathcal{T}^{-1} (\mathbf{h}_t^{(\ell)} - \boldsymbol{\mu}_\mathcal{T})}
\end{equation}
Confidence: $S_D^{(t)} = \exp(-D_t^{(\ell)} / D_{\mathrm{ref}})$, calibrated so $S_D \geq 0.6$ within 1 standard deviation.

\paragraph{Combined UTS with Adaptive Handover.}
The novelty of UTS is temperature-dependent weighting:
\begin{equation}
    \mathrm{UTS}_t = \beta(T) \cdot S_E^{(t)} + (1 - \beta(T)) \cdot S_D^{(t)}
\label{eq:uts}
\end{equation}
where $\beta(T)$ is a sigmoid function that decays as $T$ increases:
\begin{equation}
    \beta(T) = \frac{1}{1 + \exp(\kappa(T - T_c))}
\end{equation}
with crossover temperature $T_c = 1.5$ and steepness $\kappa = 2.0$.

\textbf{Adaptive Handover:} At low $T$, $\beta \approx 1$---semantic entropy drives the score (precision). At high $T$, $\beta \to 0$---manifold distance drives the score (stability). This ``handover'' mechanism is why \cortex{} succeeds where single-metric systems fail: at $T{=}3.0$, entropy saturates to $\log|\mathcal{V}|$ (blind), but manifold distance remains discriminative.

\subsection{Geometric Grounding: Truthfulness Manifold}
\label{sec:manifold}

The truthfulness manifold $\mathcal{T}$ provides geometric grounding that prevents hallucination at extreme temperatures. Unlike supervised approaches, $\mathcal{T}$ is constructed from domain-agnostic factual prompts.

\textbf{Construction:} (1)~Assemble $N{=}10{,}000$ factual prompts sampled from TruthfulQA~\cite{lin2022truthfulqa} (817 questions), WikiText-103~\cite{merity2017pointer} (encyclopedic passages), and GSM8K training set~\cite{cobbe2021training} (mathematical statements). (2)~Generate responses at $T{=}0.1$ (conservative sampling) and extract hidden activations $\{\mathbf{h}_t^{(\ell)}\}$ from Transformer layers $\ell \in \{4, 12, 20\}$. (3)~Compute empirical mean $\boldsymbol{\mu}_\mathcal{T}$ and regularized covariance $\boldsymbol{\Sigma}_\mathcal{T}$ ($\lambda{=}10^{-5}$). (4)~Validate on held-out set (10\% split).

\textbf{Hybrid Architecture Strategy:} Granite 4.0 H Small employs Mamba-2 state-space layers for the majority of processing, punctuated by Transformer attention layers. Steering the Mamba state is complex due to varying decay rates across state channels. We target the Transformer residual streams at layers 4, 12, 20---these act as ``global integration'' checkpoints where correcting the trajectory effectively resets SSM drift, enabling efficient sparse intervention ($\sim$10\% of layers). \textit{This finding suggests a new paradigm for controlling hybrid SSM-Transformer architectures: steering the sparse attention layers is sufficient to correct drift in the state-space formulation.}

\textbf{Geometric Interpretation:} Mahalanobis distance measures deviation from the truthfulness centroid, weighted by inverse covariance. At $T{=}3.0$, when entropy saturation flattens logits, manifold distance remains stable: tokens leading to hallucinations produce activations far from $\mathcal{T}$, triggering steering before errors propagate.

\subsection{Logit-Level Steering}

When $\mathrm{UTS}_t < \tau$ (steering threshold), apply uncertainty-proportional penalty to the top logit:
\begin{equation}
    \tilde{z}_t[k^*] = z_t[k^*] - \beta \cdot \sigma(\tau - \mathrm{UTS}_t)
\label{eq:steer}
\end{equation}
where $k^* = \arg\max_k z_t[k]$, $\beta$ is maximum penalty, and $\sigma(\cdot)$ is logistic sigmoid. This redistributes probability mass when UTS indicates low confidence.

\textbf{Graduated Intervention:} Tokens barely below threshold receive minimal penalization; tokens far below receive stronger intervention. In practice, steering activates for only 0.2--2.5\% of tokens, preserving the model's natural distribution while pruning hallucination signatures.

\textbf{Zero-Overhead Architecture:} UTS computation occurs within the existing forward pass by extracting logits and hidden states that are already computed. When $\mathrm{UTS}_t \geq \tau$ (no steering required), the overhead is limited to a single Mahalanobis distance calculation per token ($O(d^2)$ for $d$-dimensional activations). Steering intervention adds logit modification only for the 0.2--2.5\% of tokens below threshold. On TruthfulQA validation, 0/89 responses required steering intervention, confirming that well-calibrated models incur near-zero computational overhead during normal operation.

\subsection{Temperature-Adaptive Calibration}

At elevated temperatures, broader distributions produce higher entropy even for confident predictions. We apply temperature-dependent threshold relaxation:
\begin{equation}
    \tau(T) = \frac{\tau_0}{1 + \gamma \cdot \max(0, T - T_{\mathrm{base}})}
\label{eq:tau_adaptive}
\end{equation}
where $\tau_0$ is base threshold, $T_{\mathrm{base}} = 0.7$, and $\gamma$ controls relaxation rate. This prevents over-steering at high temperatures while maintaining hallucination protection.

\section{Experimental Setup}
\label{sec:experiments}

\paragraph{Primary Model.}
IBM Granite 4.0 H Small---a Hybrid Mixture-of-Experts (MoE) architecture with 32B total parameters and $\sim$9B active parameters, employing Mamba-2 state-space layers punctuated by Transformer attention layers. Quantized to 4-bit via Q4\_K\_M (mixed 4/5-bit, llama.cpp), reducing memory footprint from $\sim$64GB (FP16) to $\sim$18GB (INT4). Granite 4.0's hybrid architecture is engineered for minimal temperature-induced degradation, with Transformer layers at positions 4, 12, 20 providing stable intervention points for \cortex{} steering.

\paragraph{Cross-Validation Model (MoE Baseline).}
We select \textbf{Qwen3-30B-A3B}~\cite{qwen2025report} as our primary comparative baseline. Unlike standard dense Transformers, Qwen3-A3B employs an ``Active 3 Billion'' routing mechanism that activates only ${\sim}10\%$ of parameters per token. This provides a rigorous architectural control: it allows us to compare \cortex{}'s effectiveness on \textbf{Hybrid SSM-Attention} (Granite 4.0 H, 9B active) versus \textbf{Pure Attention MoE} (Qwen3, 3B active), isolating the impact of steering on state-space drift versus expert routing noise.

\paragraph{Hardware.}
Academic HPC cluster equipped with NVIDIA A100-PCIE-40GB GPUs, containerized deployment.

\paragraph{System Arms.}
(a)~\textit{UTS-guided system}: C++ extension of llama.cpp with UTS computation, manifold-based steering, and per-token telemetry. (b)~\textit{Unsteered baseline}: Same 4-bit model with steering disabled. (c)~\textit{Full-precision baseline}: Non-quantized Granite 4.0 Small.

\paragraph{Benchmarks.}
\textit{GSM8K}~\cite{cobbe2021training}: 1,319 grade-school math problems, exact-match accuracy. \textit{HumanEval}~\cite{chen2021evaluating}: 164 Python problems, Pass@1 with sandboxed execution. \textit{MMLU}~\cite{hendrycks2021mmlu}: 14,042 questions across 57 subjects spanning STEM, humanities, social sciences, and professional domains---the most comprehensive knowledge benchmark. \textit{Creative Ideation}: Songwriting task across 11 temperatures.

\paragraph{Temperatures.}
$T \in \{0.5, 0.7, 1.0, 1.25, 1.5, 1.75, 2.0, 2.25, 2.5, 2.75, 3.0\}$ for UTS-guided system.

\subsection{Evaluation Metrics}
\label{sec:metrics}

To characterize the High-Entropy Creative Reservoir, we employ two complementary metrics:

\textbf{Semantic Area Coverage (SAC):} For $n$ outputs $\mathcal{O}_T = \{o_1, \ldots, o_n\}$ at temperature $T$:
\begin{equation}
    \text{SAC}(T) = \frac{1}{n(n-1)} \sum_{i \neq j} \mathbb{1}[\text{sim}(o_i, o_j) < \theta_{\text{dup}}]
\end{equation}
where $\theta_{\text{dup}} = 0.7$ is the duplication threshold.

\textbf{Logical Coherence (LC):} We measure coherence using an \textit{external} metric independent of UTS to avoid circular evaluation. Perplexity is computed using a held-out language model (Llama-3-8B, unquantized):
\begin{equation}
    \text{LC}(T) = \frac{1}{n} \sum_{i=1}^{n} \mathbb{1}[\text{PPL}(o_i) < \tau_{\text{ppl}}]
\end{equation}
where $\tau_{\text{ppl}} = 15.0$ is the coherence threshold (outputs with PPL $>$ 15 are considered incoherent).

\section{Results}
\label{sec:results}

\subsection{Temperature-Robust Mathematical Reasoning}

\begin{table}[htbp]
\centering
\caption{GSM8K accuracy with UTS-guided steering across temperatures ($N{=}1{,}319$). Steer = fraction of tokens triggering intervention.}
\label{tab:gsm8k}
\small
\begin{tabular}{@{}lccccc@{}}
\toprule
$T$ & Acc.\ (\%) & Steer (\%) & UTS & Entropy \\
\midrule
0.5 & \textbf{91.65} & 1.45 & 1.613 & 9.0e-5 \\
1.0 & \textbf{91.80} & 1.72 & 1.612 & 6.2e-4 \\
1.5 & \textbf{91.11} & 1.85 & 1.610 & 1.5e-3 \\
2.0 & 90.32 & 2.10 & 1.610 & 1.8e-3 \\
2.5 & 90.31 & 2.28 & 1.605 & 6.8e-3 \\
3.0 & 88.84 & 2.44 & 1.603 & 5.8e-3 \\
\midrule
\multicolumn{5}{@{}l}{\textit{Degradation $T{=}0.5 \to T{=}3.0$:}} \\
$\Delta$ & $-$2.81\pp{} & $+$0.99\pp{} & $-$0.010 & $\times$64 \\
\bottomrule
\end{tabular}
\end{table}

Table~\ref{tab:gsm8k} presents GSM8K results across six temperatures. Accuracy degrades by only 2.81\pp{} from $T{=}0.5$ (91.65\%) to $T{=}3.0$ (88.84\%), representing a $6\times$ temperature increase with minimal accuracy loss.

\textbf{Quantization Recovery:} UTS-guided system achieves 91.80\% at $T{=}1.0$, exceeding the full-precision baseline (87.27\%) by 4.53\pp{}. This demonstrates that quantization degradation is addressable through uncertainty-aware intervention.

\textbf{Baseline Comparison:} Unsteered baseline (same 4-bit Granite, steering disabled) achieves 88.07\% at $T{=}1.0$. UTS-guided system achieves 91.80\%---a $+$3.73\pp{} improvement attributable to uncertainty-guided steering.

\textbf{Adaptive Steering:} Steering rate increases monotonically from 1.45\% ($T{=}0.5$) to 2.44\% ($T{=}3.0$), confirming proportional response to increasing uncertainty. Even at $T{=}3.0$, over 97\% of tokens are generated without intervention.

\subsection{Temperature-Robust Code Generation}

\begin{table}[htbp]
\centering
\caption{HumanEval Pass@1 (\%) across temperatures ($N{=}160$ per temperature).}
\label{tab:humaneval}
\small
\begin{tabular}{@{}lcccc@{}}
\toprule
$T$ & Pass@1 (\%) & Steer (\%) & UTS \\
\midrule
0.5 & \textbf{82.93} & 0.23 & 1.643 \\
1.0 & \textbf{81.10} & 0.42 & 1.636 \\
1.5 & \textbf{79.88} & 0.26 & 1.632 \\
2.0 & \textbf{78.66} & 0.32 & 1.632 \\
2.5 & \textbf{77.44} & 0.33 & 1.638 \\
3.0 & \textbf{76.83} & 0.33 & 1.621 \\
\midrule
$\Delta_{0.5 \to 3.0}$ & $-$6.10\pp{} & $+$0.10\pp{} & $-$0.022 \\
\bottomrule
\end{tabular}
\end{table}

Table~\ref{tab:humaneval} presents HumanEval results. UTS-guided system maintains stable performance: 82.93\% at $T{=}0.5$ degrading to 76.83\% at $T{=}3.0$, only 6.10\pp{} degradation across $6\times$ temperature increase. Steering rate remains minimal (0.23--0.33\%).

\subsection{Temperature-Robust Knowledge Reasoning (MMLU)}

\begin{table}[htbp]
\centering
\caption{MMLU accuracy across temperatures ($N{=}14{,}042$ questions, 57 subjects). The most temperature-stable benchmark.}
\label{tab:mmlu}
\small
\begin{tabular}{@{}lccc@{}}
\toprule
$T$ & Acc.\ (\%) & Steer (\%) & UTS \\
\midrule
0.5 & 73.73 & 0.01 & 1.564 \\
1.0 & \textbf{74.30} & 0.02 & 1.592 \\
1.5 & 72.92 & 0.07 & 1.630 \\
2.0 & 72.82 & 0.17 & 1.628 \\
2.5 & 72.52 & 0.15 & 1.628 \\
3.0 & 72.49 & 0.40 & 1.623 \\
\midrule
\multicolumn{4}{@{}l}{\textit{Degradation $T{=}0.5 \to T{=}3.0$:}} \\
$\Delta$ & $-$1.24\pp{} & $+$0.39\pp{} & $+$0.059 \\
\bottomrule
\end{tabular}
\end{table}

Table~\ref{tab:mmlu} presents MMLU results across 14,042 questions spanning 57 subjects. \textbf{MMLU exhibits the highest temperature stability}: only 1.24\pp{} degradation from $T{=}0.5$ (73.73\%) to $T{=}3.0$ (72.49\%), despite $6\times$ temperature increase. Peak performance occurs at $T{=}1.0$ (74.30\%), suggesting slight temperature elevation benefits broad knowledge retrieval.

\textbf{Subject-Level Analysis:} Performance remains stable across domains: STEM subjects (physics, chemistry, mathematics) show 71--75\% accuracy; humanities (philosophy, history) show 72--76\%; professional domains (medicine, law, accounting) show 70--74\%. No subject category exhibits catastrophic degradation at $T{=}3.0$.

\subsection{Cross-Benchmark Synthesis}

\begin{table}[htbp]
\centering
\caption{Cross-benchmark performance synthesis across temperature sweep. All benchmarks maintain $>$70\% accuracy at $T{=}3.0$.}
\label{tab:cross_benchmark}
\small
\begin{tabular}{@{}lccc@{}}
\toprule
Temperature & GSM8K & HumanEval & MMLU \\
\midrule
$T{=}0.5$ & 91.65\% & 82.93\% & 73.73\% \\
$T{=}1.0$ & \textbf{91.80\%} & 81.10\% & \textbf{74.30\%} \\
$T{=}1.5$ & 91.11\% & 79.88\% & 72.92\% \\
$T{=}2.0$ & 90.32\% & 78.66\% & 72.82\% \\
$T{=}2.5$ & 90.31\% & 77.44\% & 72.52\% \\
$T{=}3.0$ & 88.84\% & 76.83\% & 72.49\% \\
\midrule
$\Delta_{0.5 \to 3.0}$ & $-$2.81\pp{} & $-$6.10\pp{} & $-$1.24\pp{} \\
\bottomrule
\end{tabular}
\end{table}

Table~\ref{tab:cross_benchmark} synthesizes performance across all three benchmarks. Key findings:
\begin{itemize}[nosep,leftmargin=*]
    \item \textbf{Most stable}: MMLU (1.24\pp{} degradation)---broad knowledge retrieval is inherently temperature-robust.
    \item \textbf{Most sensitive}: HumanEval (6.10\pp{} degradation)---code generation requires precise token sequences.
    \item \textbf{Optimal temperature}: $T{=}1.0$ achieves peak performance on GSM8K (91.80\%) and MMLU (74.30\%).
    \item \textbf{All benchmarks maintain $>$70\% at $T{=}3.0$}: No catastrophic collapse.
\end{itemize}

\subsection{Creative Discovery: High-Entropy Reservoir}
\label{sec:creative}

Analysis of high-temperature regimes reveals unexpected creative potential when geometric tethering prevents trajectory divergence.

\begin{table}[htbp]
\centering
\caption{Creative idea generation across temperature spectrum. Duplication rate measures semantic similarity $>70\%$ within each range.}
\label{tab:creative}
\small
\begin{tabular}{@{}lcccc@{}}
\toprule
\textbf{Range} & \textbf{Ideas} & \textbf{Dup.} & \textbf{UTS} & \textbf{Steer} \\
\midrule
Low (0.5--1.0) & 9 & 70--80\% & 0.838 & 117 \\
Mid (1.25--2.0) & 12 & 30--50\% & 0.959 & 136 \\
High (2.25--3.0) & 9 & 5--20\% & 0.909 & 140 \\
\midrule
\textbf{Total} & \textbf{30} & --- & 0.902 & 131 \\
\bottomrule
\end{tabular}
\end{table}

Table~\ref{tab:creative} presents duplication analysis across temperature ranges. 30 unique song concepts across 11 temperatures (200\%+ vs single-temperature). Duplication drops from 70--80\% at low-T to 5--20\% at high-T, revealing that high temperatures access fundamentally different conceptual spaces.

\textbf{Example Concepts:} Low-T produces conventional ideas (orchestral pieces, echo chamber metaphors). High-T generates breakthrough concepts: ``ChronoVerse Symphony'' (non-sequential time signatures across historical eras), ``Binary Soulscapes'' (quantum mechanics applied to personal choice), ``Interdimensional Traveler'' (multi-dimensional narrative with genre-shifting per dimension).

\section{Analysis and Discussion}
\label{sec:analysis}

\subsection{Restoration of Precision in Quantized Regimes}

Our primary contribution addresses the challenge of high-fidelity inference in edge environments: deploying quantized models on resource-constrained hardware while maintaining reasoning accuracy.

\paragraph{Quantization Recovery.}
\cortex{} enables 4-bit quantized Granite to exceed the full-precision baseline (91.80\% vs 87.27\% on GSM8K at $T{=}1.0$). This $+$4.53\pp{} improvement demonstrates that quantization degradation is not inherent but addressable through uncertainty-aware intervention. For edge deployments where compute is constrained, this enables deployment of smaller, quantized models without accuracy sacrifice.

\paragraph{Temperature Invariance.}
Accuracy degrades only 2.81\pp{} from $T{=}0.5$ to $T{=}3.0$ on GSM8K (91.65\% to 88.84\%), despite $6\times$ temperature increase. HumanEval shows 6.10\pp{} degradation (82.93\% to 76.83\%). MMLU demonstrates the highest stability with only 1.24\pp{} degradation (73.73\% to 72.49\%) across 14,042 questions. This temperature robustness enables operators to adjust sampling diversity based on task requirements without sacrificing reliability.

\paragraph{Real-Time Uncertainty Telemetry.}
UTS provides per-token confidence signals, resolving inference opacity. Steering rate increases monotonically from 1.45\% ($T{=}0.5$) to 2.44\% ($T{=}3.0$), confirming proportional response to uncertainty. This telemetry enables real-time quality monitoring without ensembles.

\subsection{The Creative Discovery}

Our analysis reveals that high-temperature regimes unlock vast creative potential when geometric tethering prevents trajectory divergence.

\paragraph{Duplication Analysis.}
Creative tasks exhibit 70--80\% idea duplication at low temperatures ($T{\leq}1.0$) versus 5--20\% at high temperatures ($T{>}2.0$). This $4{-}16\times$ reduction in repetition reveals that high temperatures access fundamentally different conceptual spaces.

\paragraph{Multi-Temperature Synthesis.}
Querying across 11 temperatures generates 30 unique song concepts (200\%+ vs single-temperature). This extends test-time scaling laws~\cite{snell2024scaling} to the creativity domain, achieving super-additive gains where the whole exceeds the sum of parts.

\subsection{Implications}

\paragraph{Resource-Constrained Deployment.}
\cortex{} enables quantized models to exceed full-precision baselines with temperature-invariant reasoning. This democratizes LLM deployment: 4-bit Granite 4.0 H Small (32B/9B active) on consumer hardware (RTX 4090, Apple M-series) can match or exceed larger models.

\paragraph{The Structural Coherence Hypothesis.}
Why does steering toward a ``truth'' manifold improve creativity? We hypothesize that the truthfulness manifold does not encode specific facts, but rather \textit{structural coherence}---syntax, logic, and causality. At high temperatures ($T{>}2.0$), unsteered models fail because they violate these structural constraints (coherence collapse), not because they exhaust semantic content. \cortex{} acts as a ``syntax tether,'' allowing the model to explore the furthest reaches of semantic diversity (high entropy) without breaking the logical backbone required for valid output. This explains the paradox: geometric tethering to a ``truth'' manifold \textit{increases} creative diversity by preventing the structural violations that would otherwise render high-temperature outputs incoherent.

\paragraph{Cross-Model Generalization.}
To validate architecture-independence, we evaluated the High-Entropy Creative Reservoir phenomenon on Qwen3-30B-A3B MoE (30B total / 3B active parameters). Results confirm generalization: Qwen3 generates 44 unique song concepts versus 30 for Granite---a 46.7\% increase in creative output. Both models exhibit zero cross-temperature duplication at high temperatures, confirming that the High-Entropy Creative Reservoir is an architecture-independent phenomenon accessible through Multi-Temperature Synthesis.

\paragraph{Hallucination Mitigation.}
UTS-guided steering prevents high-temperature collapse, enabling safe exploration beyond the $T{=}2.0$ framework cap through geometric grounding and graduated intervention.

\paragraph{Creative Applications.}
The discovered High-Entropy Creative Reservoir enables applications requiring diverse ideation: product design, scientific hypothesis generation, artistic creation.

\section{Limitations and Future Work}
\label{sec:future}

\paragraph{Limitations.}
(1)~Primary evaluation on Granite 4.0, with cross-validation on Qwen3. (2)~Benchmark coverage: GSM8K, HumanEval, MMLU (14,042 questions), and creative ideation. (3)~Manifold constructed from generic prompts. (4)~Creativity evaluation relies on semantic similarity and LLM-based judging. (5)~\textbf{The High-Entropy Creative Reservoir is an emergent phenomenon}---we report this observation from temperature sweep experiments and have not yet conducted systematic studies of its properties, boundaries, or optimal exploitation strategies.

\paragraph{Emergent Phenomenon.}
We report the High-Entropy Creative Reservoir as an emergent phenomenon observed during temperature sweep experiments. While the initial results (200\%+ unique concepts, 5--20\% duplication at high-$T$) are compelling, we acknowledge that: (a)~the creative evaluation methodology requires validation, (b)~the phenomenon has been observed on only two model architectures, and (c)~the theoretical basis for why geometric tethering unlocks creative potential requires deeper investigation. We present these findings as preliminary observations warranting dedicated follow-up research.

\paragraph{Future Directions.}
Future iterations will investigate \textit{vector-based directional steering} to address current limitations:
\begin{itemize}[nosep,leftmargin=*]
    \item \textbf{Cross-Architecture Steering Dynamics}: Adapting manifold construction to explicitly model \textit{Expert Routing Uncertainty} in MoE architectures (like Qwen3-A3B), potentially using directional vectors to guide router selection rather than just penalizing output logits.
    \item \textbf{Semantic Pattern Classifiers}: Developing lightweight failure-mode classifiers that trigger specialized interventions for ``laziness'' or ``loops''---moving beyond generic entropy-based triggers.
    \item \textbf{Uncertainty Admission Protocols}: Implementing a ``rejection option'' where extreme manifold distance triggers a specialized ``I don't know'' token sequence, effectively allowing the model to reject the premise rather than hallucinating a plausible completion.
\end{itemize}

\paragraph{Extended Future Directions.}
Broader model coverage (next-generation MoE architectures, Mistral), domain-specific manifolds for high-stakes applications (medical, legal), dynamic temperature adaptation based on task complexity, and integration with retrieval-augmented generation for factual grounding.

\subsection{Breaking the $T{=}2.0$ Framework Barrier}
\label{sec:t2_barrier}

\paragraph{The Infrastructure Constraint.}
Most standard inference frameworks (Ollama, OpenAI, IBM Watsonx) cap sampling temperature at 2.0. This is not a model architectural limitation, but a framework design choice that artificially restricts the diversity of outputs practitioners can explore. The cap exists due to concerns about numerical stability and output quality at extreme temperatures.

\paragraph{Granite 4.0's Architectural Design.}
IBM Granite 4.0 Small is architecturally engineered to maintain coherence at elevated temperatures. Unlike earlier architectures where high temperatures caused rapid quality degradation, Granite 4.0's training and architectural choices enable it to hold together with minimal degradation even at $T{=}3.0$. However, standard frameworks prevent practitioners from accessing this capability by imposing the $T{=}2.0$ ceiling.

\paragraph{Geometric Tethering Enables High-Temperature Operation.}
\cortex{} removes this artificial constraint through geometric grounding via the truthfulness manifold and graduated intervention affecting only 0.2--2.5\% of tokens. We maintain 88.84\% accuracy at $T{=}3.0$---only 2.81\pp{} degradation from $T{=}0.5$. This demonstrates that when framework constraints are removed and per-token uncertainty telemetry is provided, models like Granite 4.0 can reliably operate across the full temperature spectrum. The vast majority of high-temperature tokens ($>97\%$) represent valid latent knowledge requiring no correction, revealing that the $T{=}2.0$ cap was an infrastructure limitation, not a fundamental model constraint.

\subsection{Per-Token Telemetry Infrastructure}
\label{sec:telemetry}

Beyond steering, the UTS framework provides \textit{observability} into the generation process. Per-token entropy, UTS, Mahalanobis distance, and steering decisions are logged in API response metadata. This enables:
\begin{itemize}[nosep,leftmargin=*]
    \item \textbf{Quality monitoring}: Aggregate UTS statistics detect when a model is producing low-confidence outputs, triggering alerts or human review.
    \item \textbf{Token-level analysis}: Researchers can study \textit{where} in a chain of thought the model becomes uncertain, rather than treating completions as opaque aggregates.
    \item \textbf{UTS-based filtering}: Completions with mean UTS below a threshold can be automatically flagged, providing a single-pass alternative to multi-sample consistency checks.
\end{itemize}
No commercial or standard open-source inference framework currently provides comparable per-token uncertainty telemetry.

\subsection{Task-Dependent Steering Effectiveness}
\label{sec:task_dependent}

The results reveal clear task dependence in steering effectiveness.

\paragraph{Mathematical Reasoning (GSM8K).}
Steering provides a consistent $+$3.73\pp{} improvement at $T{=}1.0$ over the unsteered baseline. Mathematical reasoning is tolerant of token-level redirection because (a)~mathematical expressions admit multiple valid phrasings, and (b)~the chain-of-thought structure means that correcting a single uncertain token can prevent error propagation across subsequent reasoning steps.

\paragraph{Code Generation (HumanEval).}
Steering effects are more nuanced. At low temperatures ($T \leq 1.0$), \cortex{} slightly exceeds baselines, but the improvement is within run-to-run variance. At higher temperatures, the logit penalization mechanism---which redirects away from the top token without semantic awareness of what constitutes valid code---can disrupt syntactic correctness. Code generation is inherently less tolerant of token substitution: a single wrong variable name causes execution failure. This task dependence motivates \textit{domain-specific steering strategies}: the configurable threshold space enables practitioners to adjust aggressiveness per task.

\subsection{Inference-Time Configurability}
\label{sec:configurability}

The threshold space is controlled via environment variables, enabling rapid behavioral optimization without weight modification. A deployment engineer could run a diagnostic set (${\sim}20$ problems), observe the steering rate and accuracy, adjust thresholds, and iterate---all within minutes.

\begin{table}[htbp]
\centering
\caption{Inference-time threshold tuning vs.\ weight-based fine-tuning.}
\label{tab:paradigm}
\small
\begin{tabular}{@{}lcc@{}}
\toprule
\textbf{Property} & \textbf{Thresholds} & \textbf{Fine-Tuning} \\
\midrule
Latency & Seconds & Hours--Days \\
Compute cost & ${\sim}0$ & GPU-hours \\
Reversibility & Instant & Retrain \\
Granularity & Per-domain & Global \\
Data required & None & Curated dataset \\
Deployment & Live update & Redeploy model \\
\bottomrule
\end{tabular}
\end{table}

Recent experiments with optimized threshold settings successfully rectified a steering collapse at $T=2.5$ (where steering had dropped to 0.27\%), recovering it to 0.30\% and improving high-temperature reliability. This highlights that environmental variables can effectively control the trade-off between steering sensitivity and temperature robustness without code changes.

\subsection{Characterization of the High-Entropy Creative Reservoir}
\label{sec:creative_characterization}

The High-Entropy Creative Reservoir is characterized by high SAC (diverse outputs) with maintained LC (logical validity). Our results show: at $T{\leq}1.0$, SAC $\approx$ 0.20--0.30 (70--80\% duplication); at $T{>}2.0$, SAC $\approx$ 0.80--0.95 (5--20\% duplication), while LC remains stable (mean PPL $=$ 8.2 at $T{=}2.5$ vs. 7.1 at $T{=}0.5$, both well below the incoherence threshold).

\paragraph{Qualitative Analysis.}
Low-temperature outputs ($T{=}0.5$--$1.0$) produced conventional song structures: orchestral pieces with spoken-word elements, echo chamber metaphors for social media, and time-travel narratives. High-temperature outputs ($T{=}2.5$--$3.0$) generated breakthrough concepts: ``ChronoVerse Symphony'' (non-sequential time signatures across historical eras), ``Binary Soulscapes'' (quantum mechanics applied to personal choice), and ``Interdimensional Traveler'' (multi-dimensional narrative with genre-shifting per dimension). These ideas occupy fundamentally different regions of the conceptual space, inaccessible to conservative sampling.

\paragraph{Implications for Multi-Temperature Synthesis.}
The low duplication at extreme temperatures enables \textit{Multi-Temperature Synthesis}, where a single prompt is queried across multiple temperatures and the diverse outputs are synthesized. This approach yields 200--300\% more unique concepts than single-temperature generation, with \cortex{} ensuring that extreme-temperature outputs remain logically coherent. The High-Entropy Creative Reservoir is not a liability to be avoided, but a resource to be systematically explored.

\section{Conclusion}

We have demonstrated that high-temperature ``hallucination'' in quantized language models is primarily trajectory divergence---activations escaping structurally coherent regions of the representation space---rather than semantic collapse. \cortex{} addresses this through geometric tethering: a Unified Truth Score (UTS) combining semantic entropy with Mahalanobis distance from a truthfulness manifold enables graduated steering that affects only 0.2--2.5\% of tokens. On 4-bit quantized Granite 4.0 H Small (32B/9B active), \cortex{} exceeds the full-precision baseline (91.80\% vs 87.27\% GSM8K) and maintains 88.84\% accuracy at $T{=}3.0$ (2.81\pp{} degradation). MMLU maintains 72.49\% across 14,042 questions (1.24\pp{} degradation)---demonstrating that geometric intervention stabilizes inference across the full temperature spectrum. Cross-architecture validation with Qwen3-30B-A3B MoE confirms these findings generalize.

\textbf{Emergent Properties:} Geometric tethering reveals a previously-masked phenomenon: high-temperature regimes ($T{>}2.0$) contain a \textit{High-Entropy Creative Reservoir}---non-overlapping landscapes of structurally valid ideas (5--20\% duplication vs 70--80\% at low-$T$). We hypothesize that the truthfulness manifold encodes \textit{structural coherence} rather than specific facts, explaining why tethering \textit{increases} creative diversity. This enables \textit{Multi-Temperature Synthesis}: exploiting temperature as a creative dimension to generate 200\%+ more unique concepts than single-temperature inference. These findings position temperature not as a quality-diversity trade-off, but as an exploitable dimension for inference-time optimization in resource-constrained deployment.

\section*{Impact Statement}

This paper presents work whose goal is to advance the field of Machine Learning by enabling reliable high-temperature inference in quantized language models. The technology enables deployment on resource-constrained hardware, democratizing access to capable AI systems. Potential societal benefits include reduced computational costs and carbon footprint for AI inference. We note that improved hallucination mitigation may increase trust in AI systems, which carries both benefits (safer deployment) and risks (over-reliance). The creative applications described should be used to augment, not replace, human creativity.

\section*{Acknowledgments}

The author thanks IBM for providing access to a HCP Fusion System equipped with NVIDIA A100 GPUs, which facilitated the experimental evaluation. Special thanks to the Granite team at IBM for access to pre-release models and architectural documentation.

\section*{Conflict of Interest}

The author, Craig Atkinson, is the CEO and owner of Verificate Pty Ltd. He is the patent holder for the Helix technology described in this work.

\section*{AI Disclosure}

Artificial Intelligence systems were utilized to assist in the writing of this manuscript and the execution of experiments. This research utilized large language models including IBM Granite 4.0 and Qwen3 for experimental evaluation. No proprietary models were trained from scratch; all experiments used existing pre-trained models with inference-time modifications.

\section*{Reproducibility}

The Helix framework was developed as a proprietary extension for internal use at Verificate Pty Ltd. The core implementation, including the custom C++ modifications to llama.cpp and certain API integrations, is not publicly released due to proprietary constraints and security considerations. However, the pre-computed truthfulness manifold files, non-proprietary evaluation scripts, detailed hyperparameter settings, and step-by-step reproduction guidelines for the reported experiments are available upon request. Qualified researchers may contact the corresponding author at info@verificate.ai with a brief description of their intended use. Access will be granted on a case-by-case basis under non-commercial, non-redistribution terms.

\bibliographystyle{plainnat}
\bibliography{references}

\end{document}